\documentclass[10pt,twocolumn,letterpaper]{article}

\usepackage{iccv}
\usepackage{times}
\usepackage{epsfig}
\usepackage{graphicx}
\usepackage{amsmath}
\usepackage{amssymb}
\usepackage{color,xcolor}
\usepackage{bm}
\usepackage{multirow}

\usepackage{caption}
\DeclareCaptionFont{mysize}{\fontsize{8}{9.6}\selectfont}
\usepackage{sidecap}
\sidecaptionvpos{figure}{t}

\definecolor{comment-red}{rgb}{1,0,0}

\usepackage[pagebackref=true,breaklinks=true,letterpaper=true,colorlinks,bookmarks=false]{hyperref}

\iccvfinalcopy 

\def\iccvPaperID{1449} 

\ificcvfinal\fi
\begin{document}

\title{SCAN: Structure Correcting Adversarial Network \\
for Organ Segmentation in Chest X-rays}

\author{Wei Dai, Joseph Doyle, Xiaodan Liang, Hao Zhang, Nanqing Dong, Yuan Li, Eric P. Xing\\
Petuum Inc.\\
{\tt\small \{wei.dai,joe.doyle,xiaodan.,hao.zhang,nanqing.dong,christy.li,eric.xing\}@petuum.com}
}

\maketitle

\begin{abstract}
\vspace{-10pt}
Chest X-ray (CXR) is one of the most commonly prescribed medical imaging procedures, often with over 2--10x more scans than other imaging modalities such as MRI, CT scan, and PET scans. These voluminous CXR scans place significant workloads on radiologists and medical practitioners. Organ segmentation is a crucial step to obtain effective computer-aided detection on CXR. In this work, we propose Structure Correcting Adversarial Network (SCAN) to segment lung fields and the heart in CXR images. SCAN incorporates a critic network to impose on the convolutional segmentation network the structural regularities emerging from human physiology. During training, the critic network learns to discriminate between the ground truth organ annotations from the masks synthesized by the segmentation network. Through this adversarial process the critic network learns the higher order structures and guides the segmentation model to achieve realistic segmentation outcomes. Extensive experiments show that our method produces highly accurate and natural segmentation. Using only very limited training data available, our model reaches human-level performance without relying on any existing trained model or dataset. Our method also generalizes well to CXR images from a different patient population and disease profiles, surpassing the current state-of-the-art.
\end{abstract}


\vspace{-25pt}
\section{Introduction}
\begin{figure}[!tp]
	\begin{center}
		\includegraphics[width=0.4\textwidth]{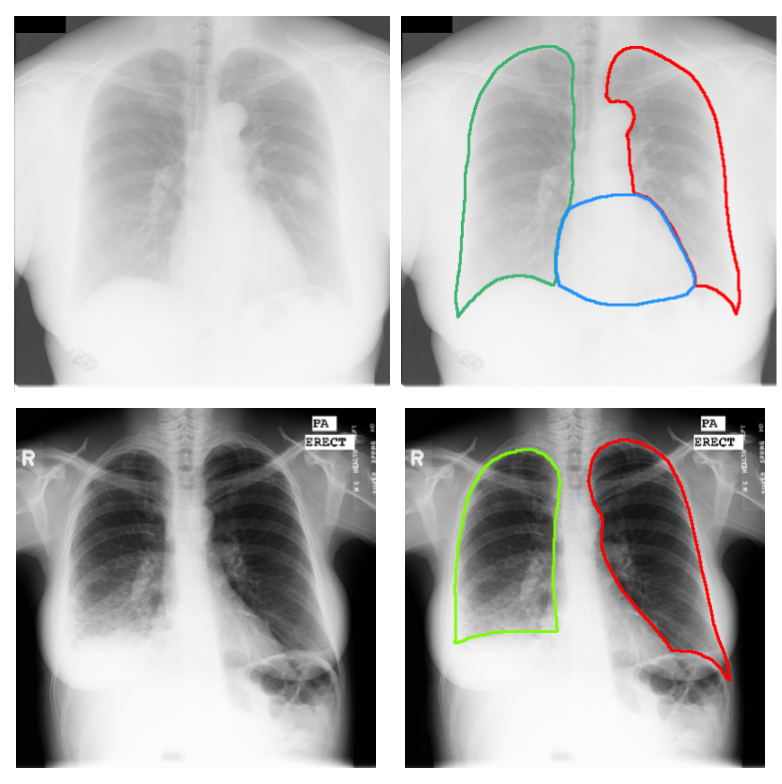}
		\vspace{-10pt}
		\caption{\small Two example chest X-ray (CXR) images from two dataset: JSRT (top) and Montgomery (bottom). From left to right columns show the original CXR images and the lung fields annotation by radiologists. JSRT (top) additionally has the heart annotation. Note that contrast can vary significantly between the dataset, and pathological lung profiles such as the bottom patient pose a significant challenge to the segmentation problem.}
		\label{fig:demo}
	\end{center}
	\vspace{-30pt}
\end{figure}

Chest X-ray (CXR) is one of the most common medical imaging modalities. 
Due to CXR's low cost and low dose of radiation, hundreds to thousands of CXRs are generated in a typical hospital daily, which creates significant diagnostic workloads. In 2015/16 year over 22.5 million X-ray images were requested in UK's public medical sector, constituting over 55\% of the total number of medical images and dominating all other imaging modalities such as computed tomography (CT) scan (4.5M) and MRI (3.1M)~\cite{nhs16}. Among X-ray images, 8 millions are Chest X-rays, which translate to thousands of CXR readings per radiologist per year. The shortage of radiologists is well documented in the developed world~\cite{shortage15,shortage14}, not to mention developing countries~\cite{tb_screen}. Compared with the more modern medical imaging technologies such as CT scan and PET scans, X-rays pose diagnostic challenges due to their low resolution and 2-D projection. It is therefore of paramount importance to develop computer-aided detection methods for chest X-rays to support clinical practitioners.


An important step in computer-aided detection on CXR images is organ segmentation. The segmentation of the lung fields and the heart provides rich structure information about shape irregularities and size measurements that can be used to directly assess certain serious clinical conditions, such as cardiomegaly (enlargement of the heart), pneumothorax (lung collapse), pleural effusion, and emphysema. Furthermore, explicit lung region masks can also improve interpretability of computer-aided detection, which is important for the clinical use.

\begin{figure}
  \centering
  \includegraphics[width=0.3\textwidth]{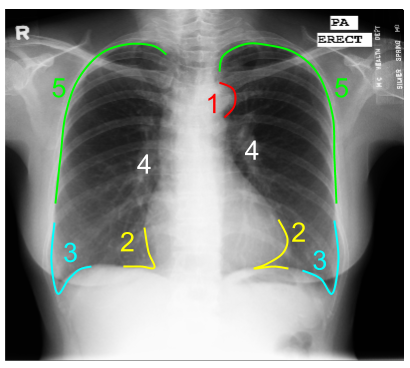} 
   \vspace{-5pt}
  \caption{Important contour landmarks around lung fields:
  aortic arch (1) is excluded from lung fields; costophrenic angles (3) and cardiodiaphragmatic angles (2) should be visible in healthy patients. Hila and other vascular structures (4) are part of the lung fields. The rib cage contour (5) should be clear in healthy lungs.}
  \label{fig:anatomy}
   \vspace{-20pt}
\end{figure}

One major challenge in CXR segmentation is to incorporate the implicit medical knowledge involved in contour determination. In the most basic sense, the positional relationship between the lung fields and the heart implies the adjacency of the lung and heart masks. Moreover, when medical experts annotate the lung fields, they look for certain consistent structures surrounding the lung fields (Fig.~\ref{fig:anatomy}). Such prior knowledge helps resolve boundaries around less clear regions caused by pathological conditions or poor imaging quality, as can be seen in Fig.~\ref{fig:demo}. Therefore, a successful segmentation model must effectively leverage global structural information to resolve the local details.

Unfortunately, unlike natural images, there is very limited CXR training data with pixel-level annotations, due to the expensive label acquisition involving medical professionals. Furthermore, CXRs exhibit substantial variations across different patient populations, pathological conditions, as well as imaging technology and operation. Finally, CXR images are gray-scale and are drastically different from natural images, which may limit the transferability of existing models. 
Existing approaches to CXR organ segmentation generally rely on hand-crafted features that can be brittle when applied on a different patient population, disease profiles, and image quality. Furthermore, these methods do not explicitly balance local information with global structure in a principled way, which is critical to achieve realistic segmentation outcomes suitable for diagnostic tasks.


In this work, we propose to use the Structure Correcting Adversarial Network (SCAN) framework that incorporates a critic network to guide the convolutional segmentation network to achieve accurate and realistic chest X-ray organ segmentation. By employing a convolutional network approach to organ segmentation, we side-step the problems faced by existing approaches based on ad hoc feature engineering. 
Our convolutional segmentation model alone can achieve performance competitive with existing methods. However, the segmentation model alone can not capture sufficient global structure to produce natural contours due to the limited training data. To impose regularization based on the physiological structures, we introduce a critic network that discriminates between the ground truth annotations from the masks synthesized by the segmentation network. The segmentation network and the critic network can be trained end-to-end. Through this adversarial process the critic network learns the higher order regularities and effectively transfers this global information back to the segmentation model to achieve realistic segmentation outcomes.





We demonstrate that SCAN produces highly realistic and accurate segmentation outcomes even when trained on very small dataset, without relying on any existing models or data from other domains. With the global structural information, our segmentation model is able to resolve difficult boundaries that require a strong prior knowledge. Using intersection-over-union (IoU) as the evaluation metric, SCAN improves the segmentation model by $1.8\%$ absolutely and achieves $94.7\%$ for the lung fields and $86.6\%$ for the heart, both of which are the new state-of-the-art by a single model, competitive with human experts ($94.6\%$ for the lungs and $87.8\%$ for the heart). We further show that SCAN model is more robust when applied to a new, unseen dataset, outperfoming the vanilla segmentation model by $4.3\%$.



\vspace{-5pt}
\section{Related Work}
\vspace{-5pt}


Our review focuses on two lines of literature most relevant to our problem: lung field segmentation and semantic segmentation with convolutional neural networks.

\noindent{\bf{Lung Field Segmentation.}} Existing work on lung field segmentation broadly falls into three categories~\cite{ginneken01}. (1) Rule-based systems apply pre-defined set of thresholding and morphological operations that are derived from heuristics~\cite{cxr_rule_based}. (2) Pixel classification methods classify the pixels as inside or outside of the lung fields based on pixel intensities~\cite{cxr_pixel01, cxr_pixel02,cxr_pixel03,cxr_pixl04}. (3) More recent methods are based on deformable models such as Active Shape Model (ASM) and Active Appearance Model~\cite{asm,aam,cxr_asm1,cxr_asm2,cxr_asm3,cxr_asm4,cxr_asm5,cxr_asm6}. Their performance can be highly variable due to the tuning parameters and whether shape model is initialized to the actual boundaries. Also, the high contrast between rib cage and lung fields can cause the model to be trapped in local minima. Our approach uses convolutional networks to perform end-to-end training from images to pixel masks without using ad hoc features. The proposed adversarial training further incorporates prior structural knowledge in a unified framework.

\begin{figure*}[!tp]
	\begin{center}
	   \vspace{-15pt}	\includegraphics[width=0.77\textwidth]{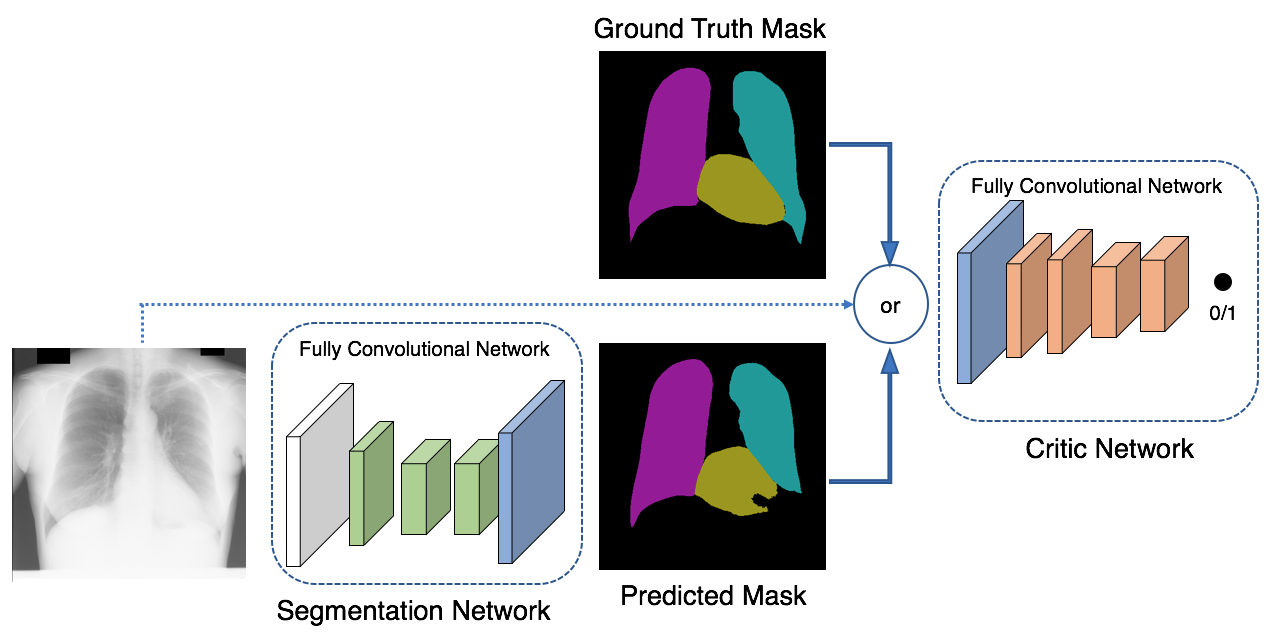}
		\vspace{-10pt}
		\caption{\small Overview of the proposed SCAN framework that jointly trains a segmentation network and a critic network with an adversarial mechanism. The segmentation network produces per-pixel class prediction. The critic takes either the ground truth label or the prediction by the segmentation network, optionally with the CXR image, and output the probability estimate of whether the input is the ground truth (with training target 1) or the segmentation network prediction (with training target 0).}
		\label{fig:overview}
	\end{center}
	\vspace{-20pt}
\end{figure*}

The current state-of-the-art method for lung field segmentation uses registration-based approach~\cite{sema}. To build a lung model for a test patient, \cite{sema} finds patients in an existing database that are most similar to the test patient and perform linear deformation of their lung profiles based on key point matching. This approach relies on the test patients being well modeled by the existing lung profiles and correctly matched key points, both of which can be brittle on a different population.



\noindent{\bf{Semantic Segmentation with Convolutional Networks.}} Semantic segmentation aims to assign a pre-defined class to each pixel, which requires a high level visual understanding. Current state-of-the-art methods for semantic segmentation use fully convolutional network (FCN)~\cite{fcn, frontend_segmentation, deeplab_crf,piecewise_fcn}. Recently~\cite{adversarial_seg} applies adversarial training to semantic segmentation and observe some improvement. These works address the natural images with color input, and are pre-trained with models such as VGG network~\cite{vgg} incorporating the learning from large-scale image classification~\cite{imagenet}. We adapt FCNs to gray-scale CXR images under the stringent constraint of a very limited training dataset of 247 images. Our FCN departs from the usual VGG architecture and can be trained without transfer learning from existing models or dataset.

Separately, U-net~\cite{u-net} and similar architectures are popular convolutional networks for biomedical segmentation and have been applied to neuronal structure~\cite{u-net} and histology images~\cite{contour-aware}. In this work we propose to use adversarial training on existing segmentation networks to enhance the global consistency of the segmentation outcomes.

We note that there is a growing body of recent works that apply neural networks end-to-end on CXR images~\cite{read_cxr,attention_cxr}. These models directly output clinical targets such as disease labels without well-defined intermediate outputs to aid interpretability. Furthermore, they generally require a large number of CXR images for training, which is not readily available for many clinical tasks involving CXR images.


\vspace{-5pt}
\section{Problem Definition}
\vspace{-5pt}

We address the problem of segmenting the left lung field, the right lung field, and the heart on chest X-rays (CXRs) in the posteroanterior (PA) view, in which the radiation passes through the patient from the back to the front. Due to the fact that CXR is a 2D projection of a 3D structure, organs overlap significantly and one has to be careful in defining the lung fields. We adopt the definition from~\cite{ginneken}: lung fields consist of all the pixels for which radiation passes through the lung but not through the following structures: the heart, the mediastinum (the opaque region between the two lungs), below the diaphragm, the aorta, and, if visible, the superior vena cava (Fig.~\ref{fig:anatomy}). The heart boundary is generally visible on two sides, while the top and bottom borders of the heart have to be inferred due to occlusion by the mediastinum. As can be seen in Fig.~\ref{fig:demo}, this definition captures the common notion of lung fields and the heart, and include regions pertinent to CXR reading in the clinical settings. 

\vspace{-5pt}
\section{Structure Correcting Adversarial Network}
\vspace{-5pt}

We detail our approach to semantic segmentation of lung fields and the heart using the proposed Structure Correcting Adversarial Network (SCAN) framework. To tailor to the special problem setting of CXR images, we develop our network architecture from ground up following the best practices and extensive experimentation. Using a dataset over an order of magnitude smaller than common semantic segmentation datasets for natural images, our model can be trained end-to-end from scratch to an excellent generalization capability without relying on existing models or datasets.


\subsection{Adversarial Training for Semantic Segmentation}

Adversarial training was first proposed in Generative Adversarial Network (GAN)~\cite{gan} in the context of generative modeling\footnote{We point out that GAN bears resemblance to the actor-critic model in the existing reinforcement learning paradigm.}. The GAN framework consists of a generator network and a critic network that engage in an adversarial two-player game, in which the generator aims to learn the data distribution and the critic estimates the probability that a sample comes from the training data instead of synthesized by the generator. The generator's objective is to maximize the probability that the critic makes a mistake, while the critic is optimized to minimize the chance of mistake. It has been demonstrated that the generator produces samples (e.g., images) that are highly realistic~\cite{dcgan15}.

A key insight in this adversarial process is that the critic, which itself can be a complex neural network, can learn to exploit higher order inconsistencies in the samples synthesized by the generator. Through the interplay of the generator and the critic, the critic can guide the generator to produce samples more consistent with higher order structures in the training samples, resulting in a more ``realistic'' data generation process.

The higher order consistency enforced by the critic is particularly desirable for CXR segmentation. Human anatomy, though exhibiting substantial variations across individuals, generally maintains stable relationship between physiological structures (Fig.~\ref{fig:anatomy}). CXRs also pose consistent views of these structures thanks to the standardized imaging procedures. We can, therefore, expect the critic to learn these higher order structures and guide the segmentation network to generate masks more consistent with the learned global structures.

We propose to use adversarial training for segmenting CXR images. Fig.~\ref{fig:overview} shows the overall SCAN framework in incorporating adversarial process to the semantic segmentation. The framework consists of a segmentation network and a critic network that are jointly trained. The segmentation network makes pixel-level predictions of the target classes, playing the role of the generator in GAN but conditioned on an input image. On the other hand, the critic network takes as input the segmentation masks and outputs the probability that the input mask is the ground truth annotations instead of the prediction by the segmentation network. The network can be trained jointly through a minimax scheme that alternates between optimizing the segmentation network and the critic network.

\vspace{-5pt}
\subsection{Training Objectives}

Let $S$, $D$ be the segmentation network and the critic network, respectively. The data consist of the input images $\bm{x}_i$ and the associated mask labels $\bm{y}_i$, where $\bm{x}_i$ is of shape $[H, W, 1]$ for a single-channel gray-scale image with height $H$ and width $W$, and $\bm{y}_i$ is of shape $[H,W,C]$ where $C$ is the number of classes including the background. Note that for each pixel location $(j,k)$, $y_i^{jkc}=1$ for the labeled class channel $c$ while the rest of the channels are zero ($y_i^{jkc'}=0$ for $c'\ne c$). We use $S(\bm{x}) \in [0,1]^{[H,W,C]}$ to denote the class probabilities predicted by $S$ at each pixel location such that the class probabilities normalize to 1 at each pixel. Let $D(\bm{x}_i, \bm{y})$ be the scalar probability estimate of $\bm{y}$ coming from the training data (ground truth) $\bm{y}_i$ instead of the predicted mask $S(\bm{x}_i)$. We define the optimization problem as
\begin{equation}
\footnotesize
\begin{split}
\min_S \max_D \Bigl\{ &J(S, D) := \sum_{i=1}^N J_s(S(\bm{x}_i), \bm{y}_i) \\
&- \lambda \Big[ J_d(D(\bm{x}_i, \bm{y}_i), 1) + J_d(D(\bm{x}_i, S(\bm{x}_i)), 0) \Big]  \Bigr\}
\label{eq:objective}
\end{split}
\end{equation}
, where $J_s(\bm{\hat{y}},\bm{y}) := \frac{1}{HW}\sum_{j,k}\sum_{c=1}^C -y^{jkc}\ln y^{jkc}$ is the multi-class cross-entropy loss for predicted mask $\bm{\hat{y}}$ averaged over all pixels. $J_d(\hat{t}, t) := -t \ln \hat{t} + (1-t)\ln(1-\hat{t})$ is the binary logistic loss for the critic's prediction. $\lambda$ is a tuning parameter balancing pixel-wise loss and the adversarial loss. We can solve Eq.~\eqref{eq:objective} by alternate between optimizing $S$ and optimizing $D$ using their respective loss functions.

\noindent{\bf{Training the Critic: }} Since the first term in Eq.~\eqref{eq:objective} does not depend on $D$, we can train our critic network by {\it minimizing} the following objective with respect to $D$ for a fixed $S$:
\begin{equation*}
\footnotesize
\sum_{i=1}^N J_d(D(\bm{x}_i, \bm{y}_i), 1) + J_d(D(\bm{x}_i, S(\bm{x}_i)), 0)
\end{equation*}

\begin{figure*}[t!]
	\begin{center}
	    \includegraphics[width=\textwidth]{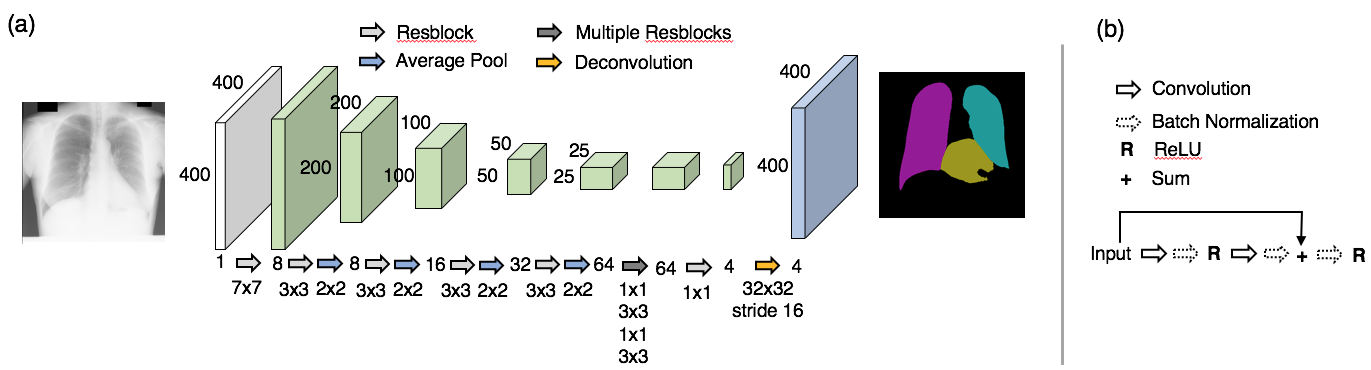}
		\vspace{-0.6cm}
		\caption{\small The segmentation network architecture. (a) Fully convolutional network for dense prediction. The feature map resolutions (e.g., 400$\times$400) are denoted only for layers with resolutions different from the previous layer. The arrows denotes the forward pass, and the integer sequence (1, 8, 16,...) is the number of feature maps. $k \times k$ below Resblock (residual block), average pool, and deconvolution arrows indicates the receptive field sizes. The dark gray arrow denotes 5 resblocks. All convolutional layers have stride $1\times 1$, while all average pooling layers have stride $2 \times 2$.  The output is the class distribution for 4 classes (3 foreground + 1 background). The total number of parameters is 271k, $\sim500$x smaller than the VGG-based down-sampling path in~\cite{fcn}. (b) The residual block architecture is based on~\cite{resblock}. The residual block maintains the same number of feature maps and spatial resolution which we omit here. Best viewed in color.}
		\label{fig:generator}
	\end{center}
	\vspace{-20pt}
\end{figure*}

\noindent{\bf{Training the Segmentation Network:}} Given a fixed $D$, we train the segmentation network by minimizing the following objective with respect to $S$:
\begin{equation*}
\footnotesize
\sum_{i=1}^N J_s(S(\bm{x}_i), \bm{y}_i) + \lambda J_d(D(\bm{x}_i, S(\bm{x}_i)), 1)
\end{equation*}
Note that we use $J_d(D(\bm{x}_i, S(\bm{x}_i)), 1)$ in place of $-J_d(D(\bm{x}_i, S(\bm{x}_i)), 0)$, following the recommendation in~\cite{gan}. This is valid as they share the same set of critical points. The reason for this substitution is that $J_d(D(\bm{x}_i, S(\bm{x}_i)), 0)$ leads to weaker gradient signals when $D$ makes accurate predictions, such as during the early stage of training.

\subsection{Segmentation Network}

Our segmentation network is a Fully Convolutional Network (FCN), which is also the core component in many state-of-the-art semantic segmentation models~\cite{fcn,frontend_segmentation,deeplab_crf}. The success of FCN can be largely attributed to convolutional neural network's excellent ability to extract high level representations suitable for dense classification. FCN can be divided into two modules: the down-sampling path and the up-sampling path. The down-sampling path consists of convolutional layers and max or average pooling layers, with architecture similar to those used in image classification~\cite{vgg}.
The down-sampling path can extract the high level semantic information, usually at a lower spatial resolution. The up-sampling path consists of convolutional and deconvolutional layers (also called transposed convolution) to predict scores for each classes at the pixel level using the output of the down-sampling path.

Most FCNs are applied to color images with RGB channels, and their down-sampling paths are initialized with parameters trained in large-scale image classification~\cite{fcn}. However, CXR is gray-scale and thus the large model capacity used in image classification networks that leverages the richer RGB input is likely to be counter-productive for our purpose. 
Furthermore, our FCN architecture has to be highly parsimonious to take into account that our training dataset of 247 CXR images is orders of magnitude smaller than those in the natural image domains. Lastly, in our task we focus on segmenting three classes (the left lung, the right lung, and the heart), which is a smaller classification space compared with dataset such as PASCAL VOC which has 20 class objects. A more parsimonious model configuration is therefore highly favorable in this setting.

Figure \ref{fig:generator} shows our FCN architecture. We find that it is advantageous to use much fewer feature maps than the conventional VGG-based down-sampling path. Specifically, we start with just 8 feature maps in the first layers, compared with 64 feature maps in the first layer of VGG~\cite{vgg}. To obtain sufficient model capacity, we instead go deep with 20 convolutional layers. We also interleave $1\times 1$ convolution with $3 \times 3$ in the final layers to emulate the bottleneck design~\cite{resnet}. All in all the segmentation network contains 271k parameters, 500x smaller than VGG-based FCN~\cite{fcn}. We employ residual blocks~\cite{resnet} (Fig.~\ref{fig:generator}(b)) to aid optimization. The parsimonious network construction allows us to optimize it efficiently without relying on any existing trained model, which is not readily available for gray-scale images.

\vspace{-5pt}
\subsection{Critic Network}
\vspace{-5pt}
\begin{figure}[!tp]
	\begin{center}
		\includegraphics[width=0.5\textwidth]{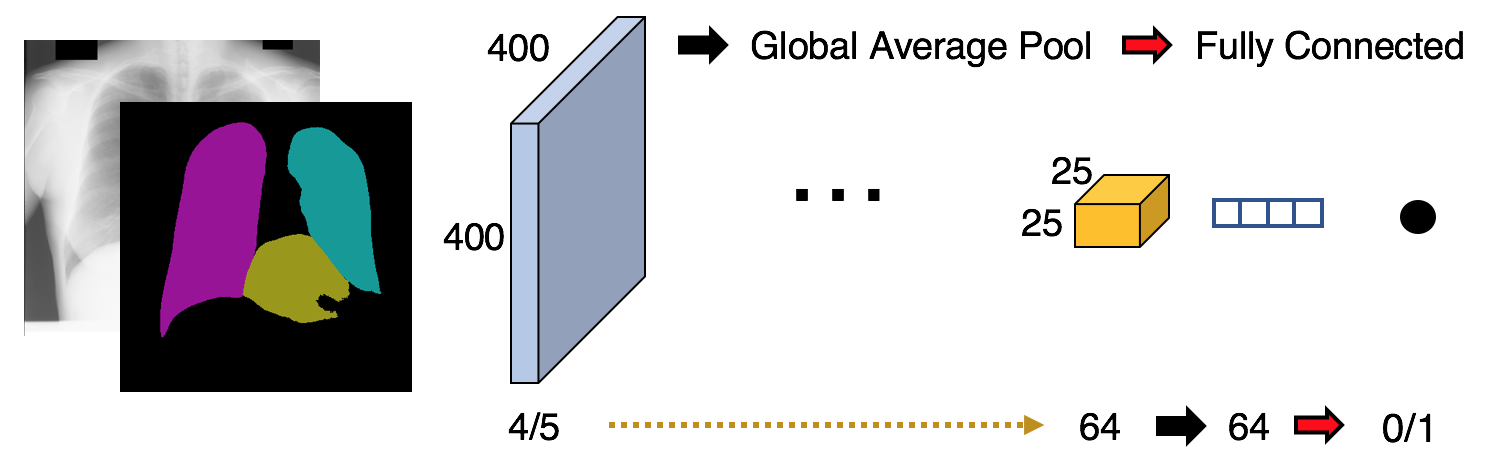}
		\caption{\small The critic network architecture. Our critic FCN mirrors the segmentation network (Fig.~\ref{fig:generator}). The input to the critic network has 4 channels or, when the input image is included, 5 channels. The subsequent layers are the same as the segmentation network up to the yellow box with 64 channel, which corresponds to the last green box in Fig.~\ref{fig:generator} with 64 channels. We omit the intermediate layers and refer readers to Fig.~\ref{fig:generator}. The training target is 0 if the input is from segmentation network, and 1 for ground truth labels.}
		\label{fig:critic}
	\end{center}
	\vspace{-0.9cm}
\end{figure}

Our critic network mirrors the construction of the segmentation network, and is also a fully convolutional network. Fig.~\ref{fig:critic} shows the architecture, omitting the intermediate layers that are identical to the segmentation networks. This way the critic network contains similar model capacity as the segmentation network with a similar field of view, which is important due to the large object size in the CXR images. We can optionally include the original CXR image as input to the critic as an additional channel, which is a more economic approach to incorporate the image in critic network than~\cite{adversarial_seg}. Preliminary experiment shows that including the original CXR image does not improve performance, and thus for simplicity we feed only the mask prediction to the critic network. Overall our critic network has 258k parameters, comparable to the segmentation network.


\vspace{-10pt}
\section{Experiments}
\vspace{-5pt}

We perform extensive evaluation of the proposed SCAN framework and demonstrate that our approach produces highly accurate and realistic segmentation of the lung fields and the heart in CXR images.

\vspace{-6pt}
\subsection{Dataset and Processing}
\vspace{-5pt}
We use two publicly available dataset to evaluate our proposed SCAN network for the segmentation of lung fields and the heart on CXR images. To the best of our knowledge, these are the only two publicly available dataset with at least lung field annotations. We point out that the dataset come from two different countries with different lung diseases, representing diverse CXR samples.

\noindent{\bf{JSRT.}} The JSRT dataset was released by Japanese Society of Radiological Technology (JSRT)~\cite{jsrt} and the lung fields and the heart masks labels were made available by~\cite{ginneken} (Fig.~\ref{fig:demo}). The dataset contains 247 CXRs, among which 154 have lung nodules and 93 have no lung nodule. All images have resolution $2048\times 2048$ in gray-scale with color depth of 12 bits. We point out that this dataset represents mostly normal lung and heart masks despite the fact that the majority of patients have lung nodule. The reason is that lung nodules in most cases do not alter the contour of the lungs and the heart, especially when the lung nodules are small.


\noindent{\bf{Montgomery.}} The Montgomery dataset contains images from the Department of Health and Human Services, Montgomery County, Maryland, USA. The dataset consists of 138 CXRs, including 80 normal patients and 58 patients with manifested tuberculosis (TB). The CXR images are 12-bit gray-scale images of dimension $4020\times 4892$ or $4892\times 4020$. Only the two lung masks annotations are available (Fig.~\ref{fig:demo}).


We scale all images to $400\times 400$ pixels, which retains sufficient visual details for vascular structures in the lung fields and the boundaries. Preliminary experiments suggests that increasing the resolution to $800\times 800$ pixels does not improve the segmentation performance, consistent with the observation in~\cite{sema}. Due to the high variation in image contrast between dataset (Fig.~\ref{fig:demo}), we perform per-image normalization. Given an image 
$\bm{x}$
we normalize it with $\tilde{x}^{jk} := \frac{x^{jk} - \bar{x}}{\sqrt{\text{var}(\bm{x})}}$, where $\bar{x}$ and $\text{var}(\bm{x})$ are the mean and variance of pixels in $\bm{x}$, respectively. Note we do not use statistics from the whole dataset. Data augmentation by rotating and zooming images, did not improve results in our preliminary experiments. We thus did not apply any data augmentation.

In post-processing, we fill in any hole in the predicted mask, and remove the small patches disjoint from the largest mask. We observe that in practice this is important for the prediction output of the segmentation network (FCN alone), but does not affect the evaluation results for FCN with adversarial training.

\vspace{-5pt}
\subsection{Training Protocols}
\vspace{-5pt}
GANs are known to be unstable during the training process and can ``collapse'' when the generator produces outcomes that lies in a much smaller subspace than the data distribution.
To mitigate this problem, we pre-train the segmentation network using only the pixel-wise loss $J_s$ (Eq.~\eqref{eq:objective}), which also gives faster training than the full adversarial training, as training the segmentation network using pixel losses involves forward and backward propagation through the segmentation network only but not the critic network. We use Adam optimizer with learning rate 0.0002 to train all models for 350 epochs, defined as a pass over the training set. We use mini-batch size 10. When training involves critic network, for each mini-batch we perform 5 optimization steps on the segmentation network for each optimization step on the critic network. Our training takes place on machines equipped with a Titan X GPU.


We use the following two metrics for evaluation: {\bf Intersection-over-Union (IoU)} is the agreement between the ground truth and the estimated segmentation mask. Formally, let $P$ be the set of pixels in the predicted segmentation mask for a class and $G$ the set of pixels in the ground truth mask for the same class. We can define IoU as $\frac{|P\cap G|}{|P\cup G|} = \frac{|TP|}{|TP| + |FP| + |FN|}$, where TP, FP, FN denotes for the set of pixels that are true positive, false positive, and false negative, respectively. {\bf Dice Coefficient} is a popular metric for segmentation in the medical domain. Using the notation defined above, Dice coefficient can be calculated as $\frac{2|P\cap G|}{|P| + |G|} = \frac{2|TP|}{2|TP| + |FP| + |FN|}$.
\vspace{-5pt}
\subsection{Experiment Design and Results}
\vspace{-5pt}

\begin{figure*}[!tp]
	\begin{center}
		\includegraphics[width=0.88\textwidth]{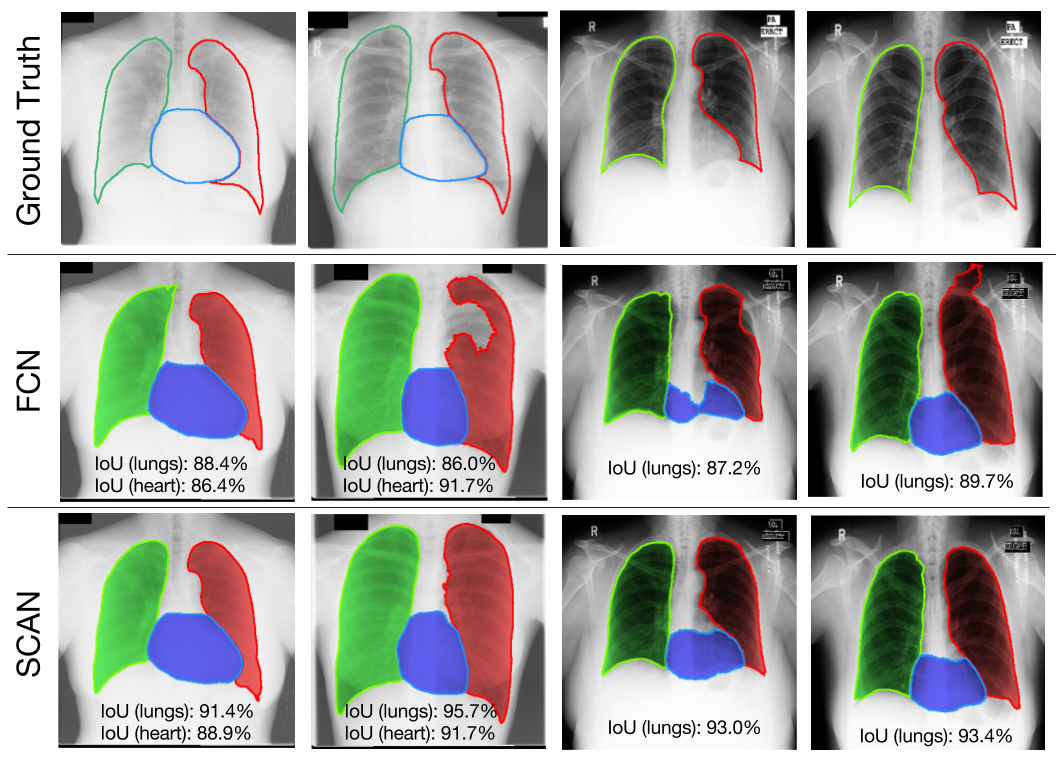}
		\vspace{-10pt}
		\caption{\small Visualization of segmentation results. From top to bottom: ground truth, FCN prediction without adversarial training, FCN prediction with adversarial training. The contours of the predicted mask are added for visual clarity. Each column is a patient. The left two columns are patients from the JSRT evaluation set with models trained on JSRT development set. The right two columns are from the Montgomery dataset using a model trained on the full JSRT dataset but not Montgomery, which is a much more challenging scenario. Note that only the two patients from JSRT dataset (left two columns) have heart annotations for evaluation of heart area IoU.}
		\label{fig:gan_vs_not}
	\end{center}
	\vspace{-8mm}
\end{figure*}

We randomly divide the JSRT dataset into the development set (209 images) and the evaluation set (38 images). We tune our architecture and hyperparameters (such as $\lambda$ in Eq.~\eqref{eq:objective}) using a validation set within the development set. Similarly, we randomly divide the Montgomery dataset into the development set (117 images) and the evaluation set (21 images). We tune our hyperparameters on the JSRT development set and use the same for the Montgomery experiments. We use FCN for the segmentation network only architecture, and SCAN for the full framework.

\textbf{Quantitative Comparison.} In our first experiment we compare FCN with SCAN when trained on the JSRT development set and tested on the JSRT evaluation set. Table~\ref{tab:gan_vs_not} shows the IoU and Dice scores. We observe that the adversarial training significantly improves the performance. In particular, IoU for the two lungs improves from 92.9\% to 94.7\%. We also find that the performance of adversarial training is robust across a range of $\lambda$: the IoU for both lungs with $\lambda=0.1,0.01,0.001$ is $94.4\%\pm 0.4\%$, $94.5\%\pm 0.4\%$, and $94.7\%\pm 0.4\%$, respectively.

\begin{table}[htbp]
\begin{center}
  \begin{tabular}{ c | c | c | c }
    \hline
    \multicolumn{2}{c|}{}& FCN & SCAN \\ \hline
    \multirow{4}{*}{IoU} & Left Lung& $91.3\% \pm 0.9\%$ & $\bm{93.8\%}\pm 0.8\%$ \\ \cline{2-4}
    & Right Lung &$94.2\% \pm 0.2\%$ & $\bm{95.5\%}\pm 0.2\%$\\ \cline{2-4}
    & Both Lungs & $92.9\%\pm 0.5\%$ & $\bm{94.7\%}\pm 0.4\%$ \\ \cline{2-4}
    & Heart &$86.5\%\pm 0.9\%$ & $86.6\%\pm 1.2\%$
    \\     \hline
    \multirow{4}{*}{Dice}
    &Left Lung& $95.4\%\pm 0.5\%$ & $\bm{96.8\%}\pm 0.5\%$ \\ \cline{2-4}
    & Right Lungs & $97.0\%\pm 0.1\%$ & $\bm{97.7\%}\pm 0.1\%$ \\ \cline{2-4}
    & Both Lungs & $96.3\%\pm 0.3\%$ & $\bm{97.3\%}\pm 0.2\%$ \\ \cline{2-4}
    & Heart & $92.7\%\pm 0.6\%$& $92.7\%\pm 0.2\%$
    \\    \hline
  \end{tabular}
\end{center}
\vspace{-0.4cm}
\caption{IoU and Dice scores on JSRT evaluation set for left lung (on the right side of the PA view CXR), right lung (on the left side of the image), both lungs, and the heart. The model is trained on the JSRT development set. }
\vspace{-0.8cm}
\label{tab:gan_vs_not}
\end{table}

In Table~\ref{tab:baselines} we compare our approach to several existing methods on the JSRT dataset, as well as human performance. Our model surpasses the current state-of-the-art method which is a registration-based model~\cite{sema} by a significant margin. Furthermore, our method is competitive with human performance for both lung fields and the heart.

\begin{table}[htbp]
\begin{center}
  \begin{tabular}{ c | c | c}
    \hline
    & IoU (Lungs) & IoU (Heart) \\ \hline
    Human Observer~\cite{ginneken} & $\bm{94.6\%}\pm 1.8\%$ & $\bm{87.8\%}\pm 5.4\%$ \\ \hline
    {\bf Ours (SCAN)} & $\bm{94.7\%}\pm 0.4\%$ & $\bm{86.6\%}\pm 1.2\%$ \\ \hline
    Registration~\cite{sema} & $92.5\%\pm 0.4\%$ & -- \\\hline
    ShRAC~\cite{ShRAC} & $90.7\%\pm 3.3\%$ & -- \\\hline
    ASM~\cite{ginneken} & $90.3\%\pm 5.7\%$ & $79.3\%\pm 11.9\%$ \\\hline
    AAM~\cite{ginneken} & $84.7\%\pm 9.5\%$ & $77.5\%\pm 13.5\%$ \\\hline
    Mean Shape~\cite{ginneken} & $71.3\%\pm 7.5\%$ & $64.3\%\pm 14.7\%$
    \\ \hline
  \end{tabular}
\end{center}
\vspace{-0.4cm}
\caption{Comparison with existing single-model approaches to lung field segmentation using JSRT dataset. Note that except the registration method, all other methods have slightly different evaluation schemes (e.g., data splits) than our evaluation. Human performance is included as a reference. IoU for heart is omitted for methods addressing only the segmentation of lung field.}
\vspace{-0.6cm}
\label{tab:baselines}
\end{table}

For clinical deployment it is important for the segmentation model to generalize to a different population with different imaging quality, such as when deployed in another country or a specialty hospital with very different disease distribution among the patients. In our next experiment we therefore train our model on the full JSRT dataset, which is collected in Japan from population with lung nodules, and test the model on the full Montgomery dataset, which is collected in the U.S. from patients potentially with TB. As can be seen in Fig.~\ref{fig:demo}, the two dataset present very different contrast and background diseases. Table~\ref{tab:montgomery} shows that FCN alone does not generalize well to a new dataset, as IoU for both lungs degrades to $87.1\%$. However, SCAN substantially improves the performance, surpassing state-of-the-art method based on registration~\cite{sema}.

\begin{figure*}[!tp]
	\begin{center}
		\includegraphics[width=0.8\textwidth]{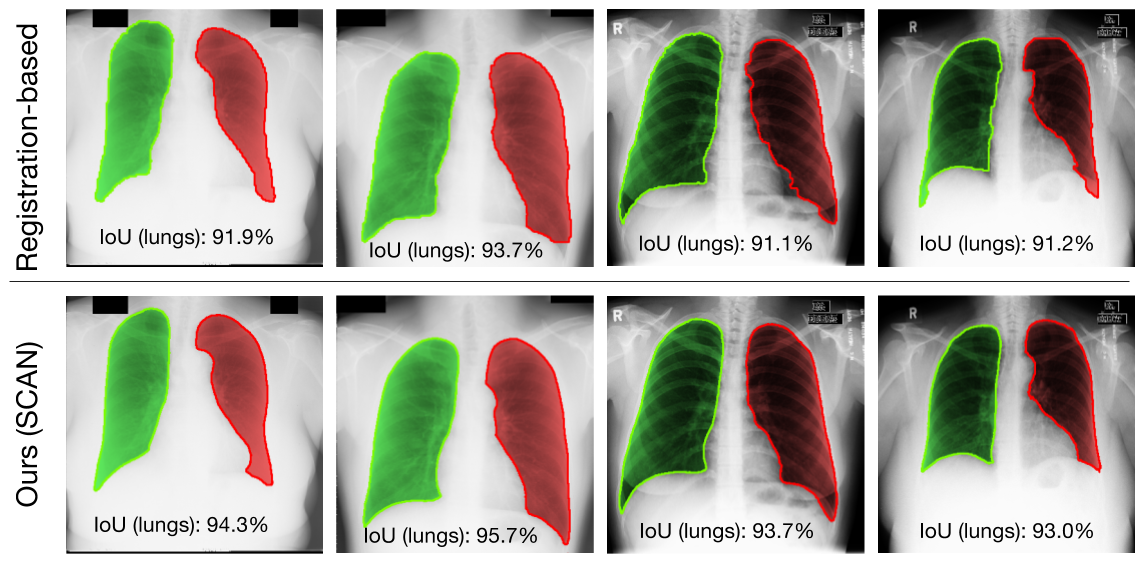}
		\vspace{-10pt}
		\caption{\small Comparison with the registration-based method that is the current state-of-the-art~\cite{sema}. The left two columns are from the JSRT evaluation set (using model trained on JSRT development set), and the right two columns are from the Montgomery set (using model trained on the full JSRT dataset). For the left two columns our method produces more realistic contours around the sharp costophrenic anglos. For the right two columns ~\cite{sema} struggles due to the mismatch between test patient lung profile and the existing lung models in JSRT.}
		\label{fig:vs_sema}
	\end{center}
	\vspace{-7mm}
\end{figure*}

We further investigate the scenario when training on the two development sets from JSRT and Montgomery {\it combined}. Without any further hyperparameter tuning, SCAN improves the IoU on two lungs to $95.1\%\pm 0.43\%$ on the JSRT evaluation set, and $93.0\%\pm 1.4\%$ on the Montgomery evaluation set, a significant improvement compared with when training on JSRT development set alone.

\textbf{Qualitative Comparison.} Fig.~\ref{fig:gan_vs_not} shows the qualitative results from these two experiments. The failure cases in the middle row by our FCN reveals the difficulties arising from CXR images' varying contrast across samples. For example, the apex of the ribcage of the rightmost patient's is mistaken as an internal rib bone, resulting in the mask ``bleeding out'' to the black background, which has similar intensity as the lung field. Vascular structures near mediastinum and anterior rib bones (which appears very faintly in the PA view CXR) within the lung field can also have similar intensity and texture as exterior boundary, leading the FCN to make the drastic mistakes seen in the middle two columns. SCAN significantly improves all of the failure cases and produces much more natural outlines of the organs. We also notice that adversarial training sharpens the segmentation of costophrenic angle (the sharp angle at the junction of ribcage and diaphragm). Costophrenic angles are important in diagnosing pleural effusion and lung hyperexpansion, among others.

\begin{table}[htbp]
\vspace{-0.2cm}
\begin{center}
  \begin{tabular}{ c | c }
    \hline
    & IoU (Both Lungs) \\ \hline
    Ours (SCAN) & $\bm{91.4\%}\pm 0.6\%$ \\ \hline
    Ours (FCN) & $87.1\%\pm 0.8\%$ \\ \hline
    Registration~\cite{sema} & $90.3\%\pm 0.5\%$ \\
    \hline
  \end{tabular}
\end{center}
\vspace{-0.5cm}
\caption{Performance on the full Montgomery dataset using model trained on the full JSRT dataset. Compared with the JSRT dataset, the Montgomery dataset exhibits much higher degree of lung abnormalities and varying imaging quality. This setting tests the robustness of the method in generalizing to a different population and imaging setting.
}
\vspace{-0.4cm}
\label{tab:montgomery}
\end{table}

Our SCAN framework is efficient at test time, as it only needs to perform a forward pass through the segmentation network but not the critic network. Table~\ref{tab:time} shows the run time of our method compared with~\cite{sema} on a laptop. \cite{sema} takes much longer due to the need to search through lung models in the training data to find similar profiles, incurring linear cost in the size of training data. In clinical setting such as TB screening~\cite{tb_screen} a fast test time result is highly desirable.

\begin{table}[htbp]
\vspace{-5pt}
\begin{center}
  \begin{tabular}{ c | c }
    \hline
    & Test time \\ \hline
    Ours (SCAN) & 0.84 seconds\\ \hline
    Registration~\cite{sema} & 26 seconds \\     \hline
    Human & $\sim$2 minutes\\ \hline
  \end{tabular}
\end{center}
\vspace{-0.5cm}
\caption{Prediction time for each CXR image (resolution $400\times 400$) from the Montgomery dataset on a laptop with Intel Core i5, along with the estimated human time.}
\vspace{-0.4cm}
\label{tab:time}
\end{table}



\vspace{-10pt}
\section{Conclusion}
\vspace{-5pt}
In this work we present the Structure Correcting Adversarial Network (SCAN) framework that applies the adversarial process to develop an accurate semantic segmentation model for segmenting the lung fields and the heart in chest X-ray (CXR) images. SCAN jointly optimizes the segmentation model based on fully convolutional network (FCN) and the adversarial critic network which discriminates the ground truth annotation from the segmentation network predictions. SCAN is simple and yet effective, producing highly accurate and realistic segmentation.
Our approach improves the state-of-the-art and achieves performance competitive with human experts.
To our knowledge this is the first successful application of convolutional neural networks to CXR image segmentation, and our method holds the promise to integrate with many downstream tasks in computer-aided detection on CXR images

\section{Acknowledgement}

We thank Carol Cheng and Ellen Sun for providing the medical insights into understanding chest x-ray and the clinical practices around it. Their help has been very instrumental to this project.



{\small
\bibliographystyle{ieee}
\bibliography{cxr}
}

\end{document}